# An Indirect Genetic Algorithm for Set Covering Problems

Uwe Aickelin, School of Computer Science, University of Nottingham, NG8 1BB UK, uxa@cs.nott.ac.uk

**Abstract**

This paper presents a new type of genetic algorithm for the set covering problem. It differs from previous evolutionary approaches first because it is an indirect algorithm, i.e. the actual solutions are found by an external decoder function. The genetic algorithm itself provides this decoder with permutations of the solution variables and other parameters. Second, it will be shown that results can be further improved by adding another indirect optimisation layer. The decoder will not directly seek out low cost solutions but instead aims for good exploitable solutions. These are then post optimised by another hill-climbing algorithm. Although seemingly more complicated, we will show that this three-stage approach has advantages in terms of solution quality, speed and adaptability to new types of problems over more direct approaches. Extensive computational results are presented and compared to the latest evolutionary and other heuristic approaches to the same data instances.

**Introduction**

In recent years, genetic algorithms have become increasingly popular for solving complex optimisation problems such as those found in the areas of scheduling or timetabling. The general approach in the past was to directly optimise problems with a genetic algorithm often coupled with a post optimisation phase, i.e. both optimisation phases are directed towards lowering the cost of solutions. The new approach presented here is different in two respects. First, a separate decoding routine, with parameters provided by the genetic algorithm, solves the actual problem. Second, the aim of this decoder optimisation is not to achieve the lowest cost solutions in the first instance. Instead, good solutions with other desirable qualities are found, which are then exploited by a post optimisation hill-climber.

Although sounding more complicated, this approach actually simplifies the algorithm into three separate components. The components themselves are relatively straightforward and self-contained making their use for other problems possible. For instance, we are currently working on a genetic algorithm to optimise the set partitioning problem. To do this the genetic algorithm module can be left intact. The decoder module will undergo only very slight changes with most changes being made in the hill-climbing module. Thus, this approach is particularly suitable for people who have expert knowledge of the problem to be optimised but who do not know evolutionary computation, because only the non-evolutionary parts have to be adjusted to the new problem.

The approach has grown out of the observation that there is no general way of including constraints into genetic algorithms. This is one of their biggest drawbacks, as it does not make them readily applicable to most real world optimisation problems. Some methods for dealing with constraints do exist, notably by use of penalty and repair functions. However, their application and success is problem specific.[1] In the modular approach, problems involving constraints are solved by the decoder module, which frees the genetic algorithm from constraint handling.

The work described in this paper has two objectives. First, to develop a fast, flexible and modular solution approach to set covering problems and second, to add to the body of knowledge on solving constrained problems using genetic algorithms indirectly. We use a modular three-stage approach. First, the genetic algorithm finds the 'best' permutation of rows and good parameters for stage two. The decoder routine, a simple heuristic, then assigns good columns to rows in given order. Finally a post-hill-climber optimises the solutions fully.

**The Set Covering Problem**

In this paper, we consider the set covering problem. This is the problem of covering the rows of an $m$-row, $n$-column, zero-one $m \times n$ matrix $a_{ij}$ by a subset of the columns at minimal cost. Formally, the problem can be defined as follows:

Defining $x_j = 1$ if column $j$ with cost $c_j$ is in the solution and $x_j = 0$ otherwise.

$$\text{Minimise} \quad \sum_{j=1}^{n} c_j x_j \tag{1}$$

$$\text{subject to} \quad \sum_{j=1}^{n} a_{ij} x_j \geq 1 \qquad \forall i \in \{1, \ldots, m\} \tag{2}$$

$$x_j, a_{ij} \in \{0,1\} \qquad \forall i \in \{1, \ldots, m\}, j \in \{1, \ldots, n\} \tag{3}$$

Constraint set (2) guarantees that each row $i$ is covered by at least one column. If the inequalities in these equations are replaced by equalities, the resulting problem is called the set partitioning problem, where over-covering is no longer allowed. The set-covering problem has been proven to be NP-complete[2] and is a model for several important applications such as crew or railway scheduling. A more detailed description of this class of problem and an overview of solution methods is given by Caprara, Fischetti and Toth.[3,4]

**Genetic Algorithms**

Genetic algorithms are generally attributed to Holland[5] and his students in the 1970s, although evolutionary computation dates back further.[6] Genetic algorithms are meta-heuristics that mimic natural evolution, in particular Darwin's idea of the survival of the fittest. Canonical genetic algorithms were not intended for function optimisation.[7] However, slightly modified versions proved very successful.[8] Many examples of successful implementations can be found in Bäck,[9] Chaiyaratana and Zalzala[10] and others.[11]

In short, genetic algorithms copy the evolutionary process by processing a population of solutions simultaneously. Starting with usually randomly created solutions better ones are more likely to be chosen for recombination with others to form new solutions; i.e. the fitter a solution, the more likely it is to pass on its information to future generations of solutions. Thus new solutions inherit good parts from old solutions and by repeating this process over many generations eventually very good to optimal solutions should be reached. In addition to recombining solutions, new solutions may be formed through mutating or randomly changing old solutions. Some of the best solutions of each generation are kept whilst the others are replaced by the newly formed solutions. The process is repeated until stopping criteria are met.

However, constrained optimisation with genetic algorithms remains difficult. The root of the problem is that simply following the building block hypothesis, i.e. combining good building blocks or partial solutions to form good full solutions, is no longer enough, as this does not check for constraint consistency. For instance, the recombination of two partially good solutions, which beforehand satisfied all constraints, might now no longer be feasible. To solve this dilemma, many ideas have been proposed,[1] however none so far has proved to be superior and widely applicable.

In this work, we will follow a new way of dealing with constraints: an indirect route. For the set covering problem this means that the genetic algorithm no longer optimises the actual problem directly. Instead, the algorithm tries to find an optimal ordering of all rows to be covered. Once such an 'optimal' ordering is found, a secondary decoding routine assigns columns following a set of rules.

The objective of this approach is the overcoming of the problem known as epistasis. For a detailed explanation of this phenomena see Reeves,[12] but briefly it describes the inability of the genetic algorithm to assemble good full solutions from promising sub-solutions due to non-linear effects between parts of different sub-solutions. For instance, in the presence of constraints sub-solutions might be feasible in their own right, but once combined they violate some constraint. Traditional genetic algorithms cannot cope with this well, and such algorithms can be deceived to search in the wrong areas.

The indirect approach presented here is one possible way of alleviating these problems. By decoupling the actual solution space from the search space of the genetic algorithm, the epistasis within the evolutionary search is reduced. Our earlier work showed that this is a promising idea to pursue with good results achieved for complex manpower scheduling and tenant selection problems.[13,14] Here, we will take these ideas one step further by introducing yet another indirect step. Rather than immediately trying to find the optimal solution through the decoder, we instead aim for good and further exploitable solutions. These are then in turn processed by a post-hill-climber. This approach is based on observations made whilst designing a traditional direct genetic algorithm for a nurse scheduling problem.[15] There it was found that by aiming for balanced solutions in the first instance and then subsequently optimising those, superior results could be achieved. Here similar observations are made.

The actual genetic algorithm component of our algorithm follows closely traditional evolutionary approaches using a permutation-based encoding. This means that a solution, as handled by the genetic algorithm, is a permuted list of all rows to be covered. For instance, if there were five rows to be covered, one possible solution might look like (2, 4, 5, 1, 3). To arrive at an actual solution to the set covering problem detailing which columns will be used, this permutation is then post-processed by the decoder routine as described in the next section.

We will use an elitist and generational approach; i.e. most of the population is eliminated at one time leaving the fittest individuals. An alternative to this would have been a steady-state approach where one or two members of the population are replaced at a time. The inherently higher convergence pressure of this approach could possibly be off-set by higher mutation rates. However, following preliminary experiments the generational approach that had worked well in the past seemed to be superior and was therefore adopted for this study. The steady-state approach will be investigated further in future research.

Fitness is measured as the cost of the target function (1). For this particular implementation, the decoder routine ensures that all solutions are feasible. For more generic approaches, this condition will not always be desirable and / or possible to be upheld as simply finding any feasible solution might be extremely difficult. In those cases a penalty function approach, penalising solutions proportional to their violation of the constraints,[16,17] seems a sensible approach. Such an approach led to success in our earlier work.[13,14]

To avoid bias from 'super-fit' individuals, which potentially could take over the population, and to enhance diversity of selection later in the search when solutions and thus fitness values will begin to converge, a rank-based selection mechanism is used. After the fitness value is evaluated for each solution, all solutions are ranked according to this measure, with ties broken by taking the member with the lower index as numbered during the initialisation. Subsequent roulette wheel type selection is based on these ranks. For instance in a population with 200 members, the fittest individual would be assigned rank 200, the second fittest rank 199 etc, down to rank 1 for the least fittest member. This results in the best individual being chosen roughly twice as often as an average individual and far more often than below average members.

Due to the nature of the permutation-based encodings, traditional crossover operators such as 1-point or uniform crossover cannot be used as they could lead to infeasible permutations. Instead, order-based crossovers have to be employed. Unless otherwise stated, the operator used in all experiments is the permutation uniform-like crossover (PUX) with p = 0.66.[13] This is a robust crossover proven for different problems and encodings. For the mutation operator a simple swap operator that exchanges the position of two genes is used. A summary of the basic indirect genetic algorithm used can be seen in Table 1.

**The Decoder**

At the heart of our approach lies the decoder function. It is this sub-algorithm that transforms the encoded permutations into actual solutions to the problem; i.e. it transforms the strings of row numbers into columns used to cover these. As outlined before, the thinking behind this is that the genetic algorithm cannot cope well with the epistasis present in the original problem. However, it can find promising regions in the simplified search space that can then be exploited by the decoder. Thus once an 'optimal' ordering has been found a relatively simple decoding routine should be able to assign the right values, i.e. once the rows have been sorted the decoder tries to find the corresponding columns.

What qualities should such a decoder therefore possess? It seems that at least, it

- Must be computationally efficient.
- Must be deterministic, i.e. the same permutation always yields the same solution.
- Must be able to reach all solutions in the solution space.
- Should produce good solutions.

Let us consider each of the above, in turn. Firstly, the decoder has to be computationally efficient, because it will be used once for each new solution. Thus, a genetic algorithm with 200 members in the population, to be run for 500 generations, will involve approximately 100,000 decoder executions. In traditional evolutionary algorithms, the fitness evaluation function is the most time-critical component. This shows a parallel between the fitness evaluation and the decoder. One could say that in many respects the decoder is an extended fitness evaluation function, since it evaluates the fitness after first decoding the solution by assigning values to the variables. It follows from the frequency of its use that the decoder must be kept computationally simple. This fits with our early conjecture that if the ordering is right, a simple decoder will achieve good solutions.

Secondly, we expect the decoder to be deterministic in the sense that the same input, i.e. a particular permutation of rows, always produces the same output, i.e. a particular set of columns. Although at first glance restricting, this characteristic is necessary to fit in with evolutionary algorithm theory. One of the foundations of the workings of these algorithms is the idea that fitter individuals will receive proportionally more samplings. However, in our case, fitness can only be measured after a string has been processed by the decoder. This fitness is then associated with the original string for reproduction purposes. If one string could decode into different solutions these would most likely have different fitness values. This could potentially confuse the genetic algorithm, as the fitness of particular solution parts is then not always the same, which is contrary to the fundamental building block hypothesis.[5]

The third criterion is that we want to be able to reach all possible solutions. This seems a sensible thing to ask at first, however on closer inspection it might not be necessary and possibly not even desirable to do so. The reason is that one strength of having a multi-level approach is that we can possibly cut down on the size of the solution space we search. For instance, high cost or infeasible solutions will generally not be in the user's interest. Therefore, we will construct a decoder that purposely biases the search towards promising regions by cutting out undesirable areas. However, by doing so we do run the risk of excluding good and possibly the optimal solution from the search. We will give an example of such a case in the next section and we will show how this possible hazard can be minimised by having intelligent weights.

Finally, we ask that the decoder is capable of producing good solutions. Clearly this is what it is designed for and should come as no surprise. However, as we will point out later, it might pay not to aim for the seemingly best solution straight away and rather go for a solution that is further exploitable by a hill-climber. As mentioned above, the decoder has to be kept computationally fast, so too complicated hill-climbing or other routines cannot be incorporated. It would often be a waste of time to apply these resource intensive hill-climbers to the whole of the population. Thus, the approach outlined in the following will be one of two stages. The decoder searches for promising feasible but not yet fully optimised solutions. These are then further exploited and optimised by a secondary decoder or hill-climber. We will show that such a two tier algorithm is superior to a single stage decoder.

**How to choose a column?**

Having established some theoretical background for the decoder, we now need to build one for the set covering problem at hand. As outlined above, the genetic algorithm is feeding permutations of the row indices into the decoder and the desired output is a low-cost list of columns used to cover all rows. For instance, an example for 5 rows and 10 columns might look as follows: The evolutionary algorithm provides the following permutation of rows: (2, 3, 4, 5, 1). This is then fed into the decoder, which assigns the following columns (4, 6, 4, 7, 1). As can be seen, the total number of columns used is less than or equal to the total number of rows to be covered as columns may cover more than one row.

To achieve this the following scheme is proposed using a methodology that ensures all solutions are feasible as every row is covered at least once. In order to build solutions, the decoder will cycle through all possible candidate columns for each row and then choose the column most suited based on the following criteria:

(1) Cost of the column. ($C_1$)

(2) How many uncovered rows does it cover? ($C_2$)

(3) How many rows does it cover in total? ($C_3$)

For each candidate column a (weighted) score $S$ will be calculated based on the above criteria, i.e. $S = w_2C_2 + w_3C_3 - w_1C_1$. The inclusion of criteria 1 and 2 is self-explanatory. Criterion 3 is included to be fairer to late-coming columns, i.e. those that cover many rows at low cost, but, because of the order of rows, are not considered in the beginning. The idea behind this is that by including such columns, others might become redundant and can then be removed. How can $C_1$, $C_2$ and $C_3$ be measured? The first and the third criteria are straightforward, as the cost information as well as the total number of rows a column covers are parameters of the model. No calculations are required at all. The second criteria needs to be calculated for each candidate by taking into account rows already covered by other columns. From all possible candidates the column with the highest score is chosen. Possible weights for the score components are discussed in the following.

The presented decoder is computationally simple as required, but does it meet all the other criteria discussed in the previous section? Encoded permutations should always decode to the same solution, provided that ties are broken in a deterministic way. For the remainder of this paper all ties will be broken on a first come first served basis. It is less clear if the decoder is capable of reaching all solutions in the solution space. Clearly all infeasible solutions have been cut out. This is in line with our discussion above, purposely biasing the search. However, one might argue that some useful information might be lost by excluding all infeasible solutions a priori, as some of them might contain good information. Based on experimental experience, it seems that on balance by excluding all infeasible solutions more is gained than by including infeasible ones that unnecessarily distract and burden the search.

However, the proposed decoder is not capable of reaching all feasible solutions either. Firstly, by choosing candidates with high scores, the search is biased towards the inclusion of low cost columns. Thus, certain columns with very high costs might never be considered at all by the decoder, which, in most cases, should be a sensible decision. Of more concern, low-cost columns might also be excluded if a certain fixed set of weights is used. Consider a row *R* that can only be covered by two columns CO1 and CO2. Further, let us assume that CO1 is cheaper than CO2 but only covers row *R* whilst CO2 covers row *R* plus an additional row. Depending on how the weights are set, either CO1 or CO2 will always be chosen to cover *R*. If, unfortunately, the weights have been set wrong, a column required for the optimal solution could be excluded.

To overcome these problems of fixed weights and to avoid having to find good weights in the first place, e.g. via complex parameter optimisation, the following extension to the algorithm is proposed. In addition to searching for the best possible permutation of row indices, the genetic algorithm will simultaneously search for good weights:[18,19] To the end of the string, as many additional genes are attached as there are weights / criteria. So, for instance, in the case of five rows and three criteria, the string will now look like this: (2, 3, 4, 5, 1; $w_1$, $w_2$, $w_3$), where $w_1$, $w_2$ and $w_3$ are the weights for the criteria as set out previously.

Originally, these weights are randomly initialised in a sufficiently large range to include suitable values. To avoid the problems of missing values and premature convergence, the weights also take place in a simple mutation operation. In every generation, there is a chance, equal to the mutation probability, that a weight is reset to a random value. In addition, these additional weight genes will not undergo normal crossover and, instead, the rank-weighted average of both parents will be passed on to the children. Thus, the weights found in better parents will dominate and eventually the population will converge to one set, or a few sets, of weights suitable to the particular set of data. Thus, more important criteria will have higher weights assigned to them.

The final component of the optimisation process is a simple hill-climbing routine employed after the decoder has finished. This is necessary because it will usually happen that rows are covered by more than one column, possibly in such a way that some columns are redundant and can be removed from the solution without influencing feasibility. The hill-climber we use will be a simple improvement routine that cycles once through all used columns in descending cost order starting with the most expensive column. If a column is found redundant, it is removed and the solution value is adjusted accordingly.

The following example will clarify the decoder's functioning. First, the genetic algorithm performs its work creating new solutions via selection, crossover and mutation. For a simple example of five rows and three criteria, one string produced might be (2, 3, 4, 5, 1; 10, 30, 15). This tells the decoder to find a suitable column for row 2, then one for row 3 etc. The score for each candidate will be calculated using a weight of 10 for the first criterion, a weight of 30 for the second and of 15 for the third. So, the decoder would start looking at all columns that can cover row 2 first. For each column $i$ that covers row 2 a score $S_i$ is then calculated as 30 x the number of uncovered rows it covers + 15 x total number of rows it covers – 10 cost of the column. For instance, if a column with a cost of 5, covers 3 rows, all of which are yet to be covered, it score would be 30 x 3 + 15 x 3 – 10 x 5 = 85.

Once the columns with the highest score have been chosen (or, in the case of a tie, the first such column), the decoder moves on to covering the next row, in this example row 3. Before moving on, the cover provided by the chosen column(s) is updated. If row 3 is already covered by the column(s) chosen so far, then the decoder proceeds to find a cover for row 4. In subsequent stages, the scores calculated for candidate columns must take into account other columns that have already been picked. This influences the second criteria, the number of uncovered rows the column covers. The decoder will then proceed in this fashion until a cover for the last row, (here, row 1) has been found. Before calculating its fitness, the solution goes through the simple hill-climber, outlined above, removing redundant columns.

Tables 2 and 3 report the results found with the above version of decoder under the 'Basic' column label. For the genetic algorithm part only one crossover operator (PUX) with p = 0.66 was used. Thus, it is a uniform-like crossover whilst at the same time showing similar proprieties as PMX regarding the number of genes retaining absolute and relative positions.[13] All experiments were carried out on a 450 MHz Pentium II PC using the freeware LCC compiler system. To compare solutions times with those of other researchers, these were adjusted in the manner suggested by Caprara, Fischettit & Toth, to DECstation 5000/240 CPU seconds.[3] This leads to some unavoidable approximations, but gives sufficient accuracy to provide some insights. The tables show the best results out of 10 runs for each data instance and the average solution time over all 10 runs with the stopping criteria as specified above.

The data used is taken From Beasley’s OR library[20] and is identical to that used in the papers our results are compared to. These comparisons are with a direct genetic algorithm labelled BeCh[21] and a Lagrangean-based heuristic labelled CFT.[3] In all, 65 data sets were used, ranging in size from 200 rows x 1000 columns to 1000 rows x 10000 columns and in density (average proportion of rows covered by a column) from 2% to 20%. Summarised results can be seen in table 2 and figure 1and detailed results in table 3. As can be seen from the tables, the results are encouraging, but weaker than those found by other researchers. In particular results are poor for the larger data sets and take very long to compute. Hence, further refinements of our strategy are required.

The experiments were also repeated for a set of fixed weights, i.e. all individuals using the same set of weights throughout the whole optimisation process rather than using adjusting weights as proposed above. The values of the fixed weights were chosen as the weight set used by the overall best individual in the final generation of each run. Thus, the weights were not necessarily the same for all ten runs on a problem set of data. The results of these experiments, which are not reported here in detail, were of significantly poorer quality. This leads us to believe that adjusting weights is superior to finding the 'best' set of weights. Finally, adjusting weights allows the algorithm to change the weights as the search progresses, for example by putting more weight on the covering criteria early in the search, and relatively higher weights on the cost criteria later on. This effect has been observed. Perhaps the biggest advantage of permitting weights to be adjusted is the fact that more variety exists within the algorithm, ultimately leading to a more thorough exploration and better solutions.

**Further Enhancements**

Although promising, the results found so far are no match for those found elsewhere in the literature for the same data sets.[3,20] In this section, we will suggest some further algorithm enhancements, namely a modification of the cost criterion, the introduction of a fourth criterion and the use of different crossover operators. The first two enhancements are intended mainly to improve the look-ahead capacity of the decoder, which currently is restricted to criterion 3, the total number of rows covered by a column. By 'look-ahead' we mean the capability of making good early choices that are not too 'greedy' and allow for equally good choices towards the end of the string. With these improvements in place, we hope our algorithm will show significantly improved results when applied to the set covering problems.

One of the most important criteria when choosing a column is its cost, as the overall aim is to arrive at a low cost solution. When comparing columns for a particular row, comparing the cost of the columns provides indeed a like-for-like comparison. However, in terms of looking ahead, simply comparing the cost of a column is not the best possible move. Consider the following example: Two columns CO1 and CO2 have only the cover of row R in common. Furthermore, both cover nine rows in total, all of which are currently uncovered. CO1 has a cost of 8 whilst CO2 has a cost of 10. Therefore, it seems at first glance that CO1 is the better choice.

One might agree that for the particular row R currently under investigation CO1 dominates CO2. However, what about the remainder of the rows they cover? It might well be that CO2 with a cost of 10 is a cheap way of covering those other eight rows, if only more expensive columns would be available to do so. Equally, it could be the case that CO1 is a cheap option for R, but an expensive one for the remainder of the rows it covers as possibly cheaper columns could do that. Hence, it seems to be sensible to use rank-based cost information rather than a direct cost-based one. To implement this, the original cost criterion $C_1$ could be split into the following two sub-criteria:

- The average column's cost rank for all uncovered rows it would cover. ($C_{1a}$)
- The average column's cost rank for all rows it would cover. ($C_{1b}$)

For example a column that covers five rows in total and whose cost amongst all columns covering each of the five rows are ranked 3rd, 5th, 3rd, 1st and 2nd would have an average all row cost rank of (3 + 5 + 3 + 1 + 2) / 5 = 2.8. The average uncovered rows cost rank would be determined in a similar fashion. The score of a candidate column is now calculated as $S = w_2C_2 + w_3C_3 - w_1(C_{1a} + C_{1b})$. Summarised results for this new type of decoder are reported in Table 2 under the 'New Cost' label. One can see that the quality of solutions has improved, although not yet to the same level as the best evolutionary approach by Beasley. The results also show the extent of the additional computational burden being introduced by these extra calculations proving how time critical any changes to the decoder routine are, especially for the larger data sets.

To improve our algorithm further, it seems we have to extend its look-ahead capabilities more. Currently, the decoder tries to get it right the first time, i.e. the decoder attempts to find the best possible solution straight away. That is, on the first (and only) pass over the permutation, the decoder has to identify the best possible match for each row. Towards the end of the string, when many columns have already been fixed, it seems a reasonable assumption that the decoder will make sensible choices for the remaining rows to be covered. However, early on in the search, even with its improved cost criterion, many choices will still be made arbitrarily. To overcome this limitation, the following fourth criterion for choosing candidate columns is introduced:

- Will the inclusion of the candidate make another column redundant? If so, what is the cost of that redundant column? ($C_4$)

Unfortunately, in this form $C_4$ is not a very viable approach because the calculations required to establish the criterion for every candidate for each row would be so computationally expensive as to dwarf the remainder of the algorithm. In the light of this, it was decided to change the criterion into the following, which should be a good indicator whilst being far less expensive computationally.

- How many rows does a column share with those already chosen? ($C_{4a}$)

Which leads to the following new equation to calculate the score of each candidate $S = w_2C_2 + w_3C_3 - w_1(C_{1a} + C_{1b}) + w_4C_{4a}$.

This new criterion in conjunction with criterion 3 (how many rows does a column cover in total) has the effect that overlapping cover is encouraged, whilst still taking into account the cost and efficiency criteria. How can this be a good thing? Because solutions will go through the hill-climber, skimming off redundant and expensive columns at negligible extra computational time. One could look at this as having introduced a second level of ‘indirectness’. The goal is not to get the best solution straight away, but instead a solution that is well balanced and can then be exploited by another simple algorithm. As the results in table 2 under the '4 Criteria' label show, this works very well here with results much improved and now very similar to the best evolutionary ones. Additionally, computation times have been significantly reduced particularly for larger data instances. This is attributed to faster convergence made possible by the synergy effects between the new criterion and the hill-climber.

The final enhancement of our algorithm will be to use a variety of crossover operators (from conservative to aggressive), controlled by the genetic algorithm. Three different order-based crossover operators are used: 1-Point equivalent crossover,[22] Partial Mapping Crossover PMX[23] and Permutation based Uniform-like Crossover PUX.[13]

When comparing the performance of these three crossover operators individually, little difference in overall solution quality was noticed (no results provided). The final results presented in tables 2 and 3 under the 'IGA' (Indirect Genetic Algorithm) label are for an algorithm where the choice of crossover is left to the evolutionary process itself. This is achieved by adding one additional gene to the string, which indicates the type of crossover used, i.e. 1 = 1-point, 2 = PUX, 3 = PMX. Originally initialised at random, during reproduction the children are created using the crossover of the fitter parent. Both children then inherit this crossover type. To avoid bias and premature convergence, the crossover parameter also undergoes mutation with the same mutation probability as the remainder of the string. A mutation in this case equals a new random initialisation.

The results show that this improves solution quality further, rivalling those of the best evolutionary approach. Moreover, average solution time is further reduced to well below that of the approach by Beasley. This can be understood by examining some optimisation runs. In the early stages of optimisation, the more aggressive crossover operators PMX and PUX dominate. However, later, when smaller changes are required, the algorithm switches to 1-point crossover and thereby cutting down on the number of generations needed for convergence and hence termination.

**Conclusions and Further Work**

This paper has presented a novel genetic algorithm approach to solving the set covering problem. The algorithm differs from previous evolutionary approaches by taking an indirect route. This is achieved by splitting the search into three distinct phases. First, the genetic algorithm finds good permutations of the rows to be covered along with suitable parameters for the second stage. The second stage consists of a decoder that builds a solution from the permutations using the parameters provided. However, the best possible solutions are not sought outright, instead good but further exploitable ones are built. Thirdly, these are then fully optimised using a hill-climber.

This approach has a number of advantages. Most importantly, the search is conducted in such a way that the genetic algorithm can concentrate on what it is best at: identifying promising regions in the solution space. Furthermore, no parameter optimisation is required as this is done by the algorithm itself automatically adjusting to requirements. By decoupling the decoder from the hill-climber, the former has a better chance of looking ahead and producing better solutions. The overall results achieved rival in quality those found by the best evolutionary algorithm whilst significantly less computation time is used.

Nevertheless, more work needs to be done in this promising area. First, the results found are not the best possible as our algorithm is outperformed by a more problem specific heuristic.[3] Our results could possibly be further enhanced by additional criteria to select the columns or a more intelligent form of mutation of the strings. Also, in light of the fast run times of our algorithm, a relaxation of the stopping criterion might improve results further. However, we feel that the strength of our approach is in its modularity and hence easy adaptability to new problems. Once a suitable order-based encoding is found, only the decoding and hill-climbing criteria need to be changed rather than having to redesign everything. We are currently investigating the use of such a slightly modified algorithm on the set partitioning problem with encouraging results.

| Parameter / Strategy | Setting |
|---|---|
| Population Size | 200 |
| Population Type | Generational |
| Initialisation | Random |
| Selection | Rank Based |
| Crossover | PUX Order-Based crossover |
| Swap Mutation Probability | 1.5% |
| Replacement Strategy | Keep 20% Best of each Generation |
| Stopping Criteria | No improvement for 50 generations |

| | **BeCh[21]** | | **CFT[3]** | | **Basic** | | **New Cost** | | **4 Criteria** | | **IGA** | |
|---|---|---|---|---|---|---|---|---|---|---|---|---|
| Problem Set | Dev. | Time | Dev. | Time | Dev. | Time | Dev. | Time | Dev. | Time | Dev. | Time |
| 4 | 0.00% | 163 | 0.00% | 6.5 | 0.22% | 33.6 | 0.00% | 112.4 | 0.00% | 76.5 | 0.00% | 93.3 |
| 5 | 0.09% | 540.2 | 0.00% | 3.2 | 0.25% | 49.5 | 0.16% | 85.4 | 0.00% | 78.6 | 0.00% | 61.2 |
| 6 | 0.00% | 57.2 | 0.00% | 9.4 | 1.38% | 66 | 0.96% | 54.3 | 0.00% | 10.2 | 0.00% | 7.6 |
| A | 0.00% | 149.4 | 0.00% | 106.6 | 0.56% | 146.4 | 0.44% | 182.4 | 0.06% | 79.2 | 0.00% | 81 |
| B | 0.00% | 155.4 | 0.00% | 7.4 | 1.06% | 337.2 | 0.94% | 232.4 | 0.00% | 104.7 | 0.00% | 30.4 |
| C | 0.00% | 199.2 | 0.00% | 66 | 0.35% | 277.2 | 0.11% | 368.22 | 0.00% | 145.3 | 0.00% | 82.8 |
| D | 0.00% | 230.4 | 0.00% | 17.2 | 2.47% | 721.2 | 2.02% | 1100.2 | 0.48% | 220.4 | 0.32% | 69 |
| E | 0.00% | 8724.2 | 0.00% | 118.2 | 0.67% | 1592.4 | 0.59% | 2395.7 | 0.00% | 120.5 | 0.00% | 56 |
| F | 0.00% | 2764.8 | 0.00% | 109 | 1.54% | 2125.2 | 0.68% | 3154.3 | 0.21% | 450.4 | 0.00% | 142.8 |
| G | 0.13% | 12851.4 | 0.00% | 504.8 | 4.70% | 2827.2 | 3.84% | 4343.2 | 0.13% | 687.4 | 0.13% | 342.8 |
| H | 0.63% | 6341.6 | 0.00% | 858.2 | 5.68% | 3188.4 | 4.55% | 4123.8 | 1.88% | 701.5 | 1.30% | 412 |
| Overall | 0.08% | 2925 | 0.00% | 164 | 1.72% | 1033 | 1.30% | 1468 | 0.25% | 243 | 0.16% | 125 |

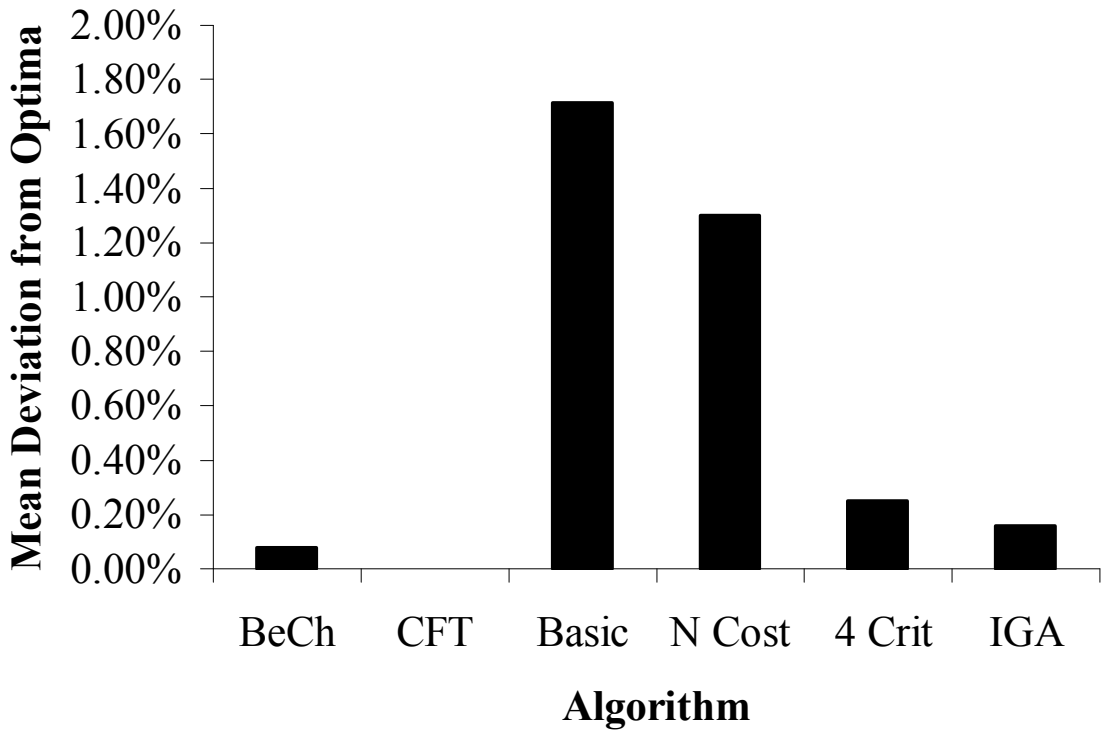

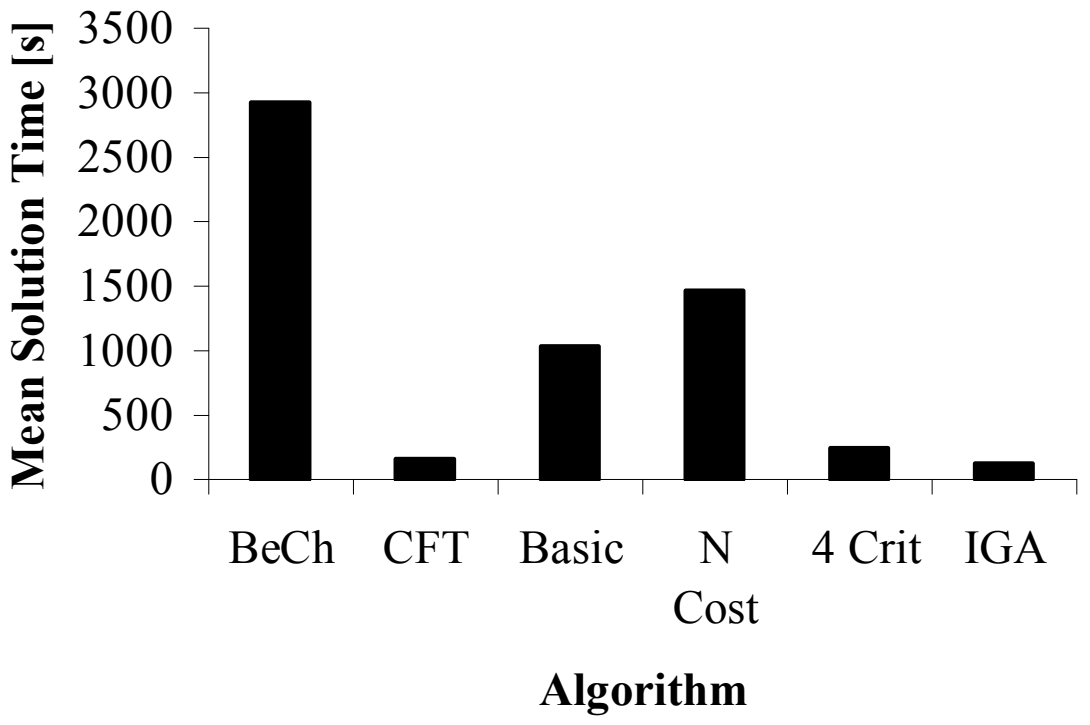

| | | | | BeCh[21] | | CFT[3] | | Basic | | IGA | |
|---|---|---|---|---|---|---|---|---|---|---|---|
| Problem | Size | Density | Optimum | Sol | Time | Sol | Time | Sol | Time | Sol | Time |
| 4.1 | 200x1000 | 2% | 429 | 429 | 295 | 429 | 2 | 429 | 36 | 429 | 105 |
| 4.2 | 200x1000 | 2% | 512 | 512 | 9 | 512 | 1 | 512 | 42 | 512 | 57 |
| 4.3 | 200x1000 | 2% | 516 | 516 | 16 | 516 | 2 | 520 | 36 | 516 | 63 |
| 4.4 | 200x1000 | 2% | 494 | 494 | 142 | 494 | 10 | 495 | 27 | 494 | 90 |
| 4.5 | 200x1000 | 2% | 512 | 512 | 44 | 512 | 2 | 512 | 18 | 512 | 120 |
| 4.6 | 200x1000 | 2% | 560 | 560 | 16 | 560 | 19 | 560 | 33 | 560 | 39 |
| 4.7 | 200x1000 | 2% | 430 | 430 | 139 | 430 | 3 | 433 | 42 | 430 | 144 |
| 4.8 | 200x1000 | 2% | 492 | 492 | 819 | 492 | 22 | 492 | 21 | 492 | 93 |
| 4.9 | 200x1000 | 2% | 641 | 641 | 136 | 641 | 2 | 644 | 24 | 641 | 159 |
| 4.10 | 200x1000 | 2% | 514 | 514 | 14 | 514 | 2 | 514 | 57 | 514 | 63 |
| 5.1 | 200x2000 | 2% | 253 | 253 | 42 | 253 | 3 | 253 | 51 | 253 | 27 |
| 5.2 | 200x2000 | 2% | 302 | 302 | 1333 | 302 | 2 | 307 | 45 | 302 | 81 |
| 5.3 | 200x2000 | 2% | 226 | 228 | 11 | 226 | 2 | 228 | 36 | 226 | 39 |
| 5.4 | 200x2000 | 2% | 242 | 242 | 10 | 242 | 2 | 242 | 27 | 242 | 120 |
| 5.5 | 200x2000 | 2% | 211 | 211 | 15 | 211 | 1 | 211 | 30 | 211 | 87 |
| 5.6 | 200x2000 | 2% | 213 | 213 | 30 | 213 | 1 | 213 | 87 | 213 | 15 |
| 5.7 | 200x2000 | 2% | 293 | 293 | 195 | 293 | 15 | 293 | 48 | 293 | 135 |
| 5.8 | 200x2000 | 2% | 288 | 288 | 3733 | 288 | 2 | 288 | 69 | 288 | 27 |
| 5.9 | 200x2000 | 2% | 279 | 279 | 14 | 279 | 3 | 279 | 78 | 279 | 48 |
| 5.10 | 200x2000 | 2% | 265 | 265 | 19 | 265 | 1 | 265 | 24 | 265 | 33 |
| 6.1 | 200x1000 | 5% | 138 | 138 | 46 | 138 | 23 | 142 | 60 | 138 | 9 |
| 6.2 | 200x1000 | 5% | 146 | 146 | 211 | 146 | 18 | 147 | 69 | 146 | 5 |
| 6.3 | 200x1000 | 5% | 145 | 145 | 12 | 145 | 2 | 148 | 60 | 145 | 3 |
| 6.4 | 200x1000 | 5% | 131 | 131 | 5 | 131 | 2 | 131 | 60 | 131 | 3 |
| 6.5 | 200x1000 | 5% | 161 | 161 | 12 | 161 | 2 | 163 | 81 | 161 | 18 |
| A.1 | 300x3000 | 2% | 253 | 253 | 222 | 253 | 82 | 255 | 180 | 253 | 105 |
| A.2 | 300x3000 | 2% | 252 | 252 | 328 | 252 | 116 | 256 | 162 | 252 | 96 |
| A.3 | 300x3000 | 2% | 232 | 232 | 127 | 232 | 250 | 233 | 147 | 232 | 51 |
| A.4 | 300x3000 | 2% | 234 | 234 | 46 | 234 | 5 | 234 | 87 | 234 | 108 |
| A.5 | 300x3000 | 2% | 236 | 236 | 24 | 236 | 80 | 236 | 156 | 236 | 45 |
| B.1 | 300x3000 | 5% | 69 | 69 | 20 | 69 | 4 | 70 | 519 | 69 | 30 |
| B.2 | 300x3000 | 5% | 76 | 76 | 12 | 76 | 6 | 77 | 252 | 76 | 27 |
| B.3 | 300x3000 | 5% | 80 | 80 | 710 | 80 | 18 | 80 | 351 | 80 | 13 |
| B.4 | 300x3000 | 5% | 79 | 79 | 30 | 79 | 6 | 81 | 405 | 79 | 78 |
| B.5 | 300x3000 | 5% | 72 | 72 | 5 | 72 | 3 | 72 | 159 | 72 | 4 |
| C.1 | 400x4000 | 2% | 227 | 227 | 188 | 227 | 74 | 227 | 312 | 227 | 132 |
| C.2 | 400x4000 | 2% | 219 | 219 | 41 | 219 | 64 | 221 | 240 | 219 | 33 |
| C.3 | 400x4000 | 2% | 243 | 243 | 541 | 243 | 70 | 245 | 420 | 243 | 171 |
| C.4 | 400x4000 | 2% | 219 | 219 | 145 | 219 | 62 | 219 | 213 | 219 | 45 |
| C.5 | 400x4000 | 2% | 215 | 215 | 81 | 215 | 60 | 215 | 201 | 215 | 33 |
| D.1 | 400x4000 | 5% | 60 | 60 | 14 | 60 | 23 | 61 | 633 | 60 | 177 |
| D.2 | 400x4000 | 5% | 66 | 66 | 199 | 66 | 22 | 66 | 498 | 66 | 51 |
| D.3 | 400x4000 | 5% | 72 | 72 | 785 | 72 | 23 | 75 | 840 | 72 | 30 |
| D.4 | 400x4000 | 5% | 62 | 62 | 74 | 62 | 8 | 63 | 1002 | 63 | 6 |
| D.5 | 400x4000 | 5% | 61 | 61 | 80 | 61 | 10 | 64 | 633 | 61 | 81 |
| E.1 | 500x5000 | 10% | 29 | 29 | 38 | 29 | 26 | 29 | 1161 | 29 | 17 |
| E.2 | 500x5000 | 10% | 30 | 30 | 14648 | 30 | 408 | 31 | 2346 | 30 | 63 |
| E.3 | 500x5000 | 10% | 27 | 27 | 28360 | 27 | 94 | 27 | 2163 | 27 | 60 |
| E.4 | 500x5000 | 10% | 28 | 28 | 540 | 28 | 26 | 28 | 1278 | 28 | 41 |
| E.5 | 500x5000 | 10% | 28 | 28 | 35 | 28 | 37 | 28 | 1014 | 28 | 99 |
| F.1 | 500x5000 | 20% | 14 | 14 | 76 | 14 | 33 | 14 | 3510 | 14 | 21 |
| F.2 | 500x5000 | 20% | 15 | 15 | 78 | 15 | 31 | 15 | 1059 | 15 | 44 |
| F.3 | 500x5000 | 20% | 14 | 14 | 267 | 14 | 249 | 14 | 1392 | 14 | 234 |
| F.4 | 500x5000 | 20% | 14 | 14 | 210 | 14 | 31 | 14 | 1863 | 14 | 174 |
| F.5 | 500x5000 | 20% | 13 | 13 | 13193 | 13 | 201 | 14 | 2802 | 13 | 241 |
| G.1 | 1000x10000 | 2% | 176 | 176 | 30200 | 176 | 147 | 182 | 3708 | 176 | 144 |
| G.2 | 1000x10000 | 2% | 154 | 155 | 361 | 154 | 783 | 161 | 2691 | 155 | 327 |
| G.3 | 1000x10000 | 2% | 166 | 166 | 7842 | 166 | 978 | 176 | 1182 | 166 | 408 |
| G.4 | 1000x10000 | 2% | 168 | 168 | 25305 | 168 | 379 | 178 | 2988 | 168 | 303 |
| G.5 | 1000x10000 | 2% | 168 | 168 | 549 | 168 | 237 | 174 | 3567 | 168 | 532 |
| H.1 | 1000x10000 | 5% | 63 | 64 | 1682 | 63 | 1451 | 66 | 3513 | 63 | 668 |
| H.2 | 1000x10000 | 5% | 63 | 64 | 530 | 63 | 887 | 67 | 3123 | 66 | 443 |
| H.3 | 1000x10000 | 5% | 59 | 59 | 1804 | 59 | 1560 | 64 | 2472 | 59 | 648 |
| H.4 | 1000x10000 | 5% | 58 | 58 | 27242 | 58 | 238 | 61 | 3018 | 59 | 235 |
| H.5 | 1000x10000 | 5% | 55 | 55 | 450 | 55 | 155 | 57 | 3816 | 55 | 66 |

Table1: Parameters and strategies used for the indirect genetic algorithm.

Table 2 Summarised results averaged for data sets of same size and density.

Table 3: Detailed Results for the Set Covering Problem

Figure 1: Graphical comparison of different heuristic approaches to the set covering problem.

# An Indirect Genetic Algorithm for Set Covering Problems

## Contribution Statement

The Set Covering Problem (SCP) is a main model for many important applications in the field of Operational Research. For instance, it can be used to describe staff scheduling, railway timetabling or aircraft scheduling type problems. The SCP has been proven to be NP-hard, therefore to solve problems of the size encountered in practice heuristic algorithms have to be employed.

Amongst these heuristic methods, evolutionary algorithms and in particular genetic algorithms have become increasingly popular in recent years. These algorithms have been successfully used to solve NP-hard problems, amongst them the SCP. However often these algorithms were only able to achieve this by employing problem specific strategies and therefore lacked the flexibility and robustness to be used for different problems.

The approach presented in this paper is a new type of self-tuning genetic algorithm. This new algorithm solves the problem indirectly. This has the advantage that the genetic algorithm component is almost independent from the problem specific decoder component. Hence, it can be re-used for other problems unchanged. Thus, our approach could be used by users unfamiliar with evolutionary computation.

Furthermore, we also explain that the decoder itself can remain intact with only minor problem specific modifications required for different problems. As will be seen, the success of this approach lies in not seeking 'optimal' solution directly, but by gradually improving solutions over a number of optimisation steps. Extensive computational results are presented and compare favourably to solutions found by the latest evolutionary and other heuristic approaches to the same data instances.